\documentclass[10pt,journal,compsoc]{IEEEtran}
\usepackage{siunitx,algorithm,graphicx,amsmath, amssymb,url}
\usepackage[noend]{algpseudocode}
\usepackage[nocompress]{cite}
\usepackage[english]{babel}

\makeatletter
\long\def\@IEEEtitleabstractindextextbox#1{\parbox{0.922\textwidth}{#1}}
\makeatother

\makeatletter
\def\BState{\State\hskip-\ALG@thistlm}
\makeatother

\newcommand{\vir}[1]{``#1"}
\newcommand*{\vv}[1]{\overrightarrow{\mkern0mu#1}}
\begin{document}
	\title{Learning from Data to Speed-up Sorted Table Search Procedures: Methodology  and Practical Guidelines} 
	
	\author{Domenico~Amato,	Raffaele~Giancarlo, Giosu\'e~Lo~Bosco
		
		\IEEEcompsocitemizethanks{
			\IEEEcompsocthanksitem
			This research is funded in part by MIUR Project of National Relevance 2017WR7SHH “Multicriteria Data Structures and Algorithms: from compressed to learned indexes, and beyond”. We also acknowledge an NVIDIA Higher Education and Research Grant (donation of a Titan V GPU). Additional support to RG has been granted by Project INdAM - GNCS  “Analysis and Processing of Big Data based on Graph Models”.\protect\\
			
			\IEEEcompsocthanksitem D. Amato,  R. Giancarlo and G.Lo Bosco are with the Department of Mathematics and Computer Science, University of Palermo, Palermo,	Italy.\protect\\
			\IEEEcompsocthanksitem G. Lo Bosco is the corresponding author,
			E-mail: giosue.lobosco@unipa.it}}
	\IEEEtitleabstractindextext{%
		\begin{abstract}
			Sorted Table Search Procedures are the quintessential query-answering tool, with widespread usage that now includes also Web Applications, e.g, Search Engines (Google Chrome)  and ad Bidding Systems (AppNexus). Speeding them up, at very little cost in space, is still a quite significant achievement. Here we study to what extend Machine Learning Techniques can contribute to obtain such a speed-up via a systematic experimental comparison of known efficient implementations of Sorted Table Search procedures, with different Data Layouts,   and their Learned counterparts developed here. We characterize the scenarios in which those latter can be profitably used with respect to the former, accounting for both CPU and GPU computing. Our approach contributes also to the study of Learned Data Structures, a recent proposal to improve the time/space performance of fundamental Data Structures, e.g., B-trees, Hash Tables, Bloom Filters. Indeed,  we also formalize an Algorithmic Paradigm of Learned Dichotomic Sorted Table Search procedures that naturally complements the Learned one proposed here and that characterizes most of the known Sorted Table Search Procedures as having a “learning phase” that approximates Simple Linear Regression.
		\end{abstract}
		
		}

	\maketitle
	
	\IEEEdisplaynontitleabstractindextext
	\IEEEpeerreviewmaketitle
	\IEEEraisesectionheading{\section{Introduction}\label{sec:introduction}}

	\IEEEPARstart{S} orted Table Search Procedures are a fundamental part of Computer Science \cite{KnuthS}. As well pointed out by 
	Khoug and Morin \cite{Morin17}, in view of the novel hardware architectures now available, a speed-up of those procedures it is still worth of investigation and useful in many application domains, e.g. Decision Support in Main Memory \cite{rao1999cache},   Web Search Engines and ad Bidding \cite{Morin17}.  
	
	An emerging trend, in order to obtain time/space improvements in classic Data Structures, is to combine Machine Learning techniques with algorithmic techniques. This new area goes under the name of Learned Data Structures \cite{kraska2018case}.  The State of the Art is well presented in a recent review by Ferragina and Vinciguerra \cite{Ferragina:2020book}.  The theme common to those new approaches to Data Structures Design and Engineering is that a query to a data structure is either intermixed with or preceded by a query to a Classifier \cite{duda20} or a Regression Model \cite{FreedmanStat}, those two being the learned part of the data structure.  
	
	Somewhat surprisingly, the rather simple Sorted Table Search Procedures, i.e., Binary \cite{KnuthS} and Interpolation Search \cite{Peterson57}, have not been considered in this new scenario. Indeed, a rather delicate balance between the power of a learning method and the speed of those routines must be thoroughly investigated. 
	We anticipate that the hearth of this research regards this balance.   Informally,  a Learned Sorted Table Search procedure consists of a learning phase, in which the CDF of the data distribution that has generated the table is learned via the elements present in the table with the use of a  Regression Model, e..g., a regression line. Then, given a query, the model is used to predict an interval in the table where the query element must be searched for.  We provide a systematic experimental analysis of the following aspects:  (1) which model complexities, e.g., Neural Networks (NN for short) or \vir{closed form} Linear Regression,  can be afforded in order for Learned Sorted Search procedures to be competitive with respect to their standard counterparts;  (2) how learnable must be a table for Learned Sorted Search procedures to be profitable;  and finally (3) when GPU computing can be afforded for query batch processing, assess to what extend are Learned Sorted Search procedures worth of consideration both when implemented in CPU and GPU. 
	
	This latter aspect is investigated since one of the motivations for the use of Machine Learning to improve data structures performance is given by the perceived advantage of Tensor computing as opposed to the classic \vir{if-then-else} approach.
	Apart from the practical indications that our experimental study provides, necessarily bound to the hardware architectures we use, it also sheds light on methodological issues regarding Learned Data Structures that have been either sketched in previous work, i.e., point (1), or largely ignored, i.e.,  (2) and (3).  Details on our experimental methodology and findings follow.
	
	Our study considers several versions of Binary \cite{KnuthS}   and Interpolation Search \cite{Peterson57}, together with efficient data layouts. They are presented in Section \ref{STSP}. Following Kraska et al., we cast the problem of Sorted Table Search in terms of  Regression Analysis \cite{FreedmanStat}, the novelty with respect to the State of the Art \cite{Ferragina:2020book} being the formalization of an Algorithmic  Paradigm, here referred to as 
	Learned Dichotomic Sorted Table Search procedures. In that paradigm, standard procedures have a \vir{learning phase} that is an approximation of Simple Linear Regression. The paradigm, mostly of methodological interest, makes explicit an overlooked connection between Learned and standard Sorted Table Search procedures, accounting also, for some recent heuristics that have been proposed. This part is in Section \ref{sec:LST}. For our experiments, we consider benchmark datasets that have appeared in the Literature and in order to conduct an in-depth investigation of Learned Sorted Table Search procedures,  we also consider synthetic datasets generated from three distribution, whose CDF is representative of the CDFs associated to the major distributions considered for Data Analysis and Modelling. That is, our synthetic datasets well represent the spectrum of CDFs that a model for Learned Table Search procedures must be asked to learn. This part is in Section  \ref{sec:Datasets}. Our experiments are reported in Sections \ref{sec:exp1} and  \ref{sec:convenient}. Based on them, a synopsis of our additional findings is given next.
	
	When no batch query processing can be afforded, or a GPU is not available, Learned Binary Search can be profitably used, provided that (a) the data is easy to learn via a very fast procedure such as Simple Linear Regression; (b) the table fits in the cache memory hierarchy. When batch processing can be afforded and a GPU is available, Learned Sorted Table  Search procedures  (both in GPU and CPU) are not competitive with respect to a simple-minded GPU Binary Search. Interestingly, Learned Interpolation Search is consistently superior to Interpolation Search,  but with  no  practical effect, since Binary Search with an 
	Eytzinger array layout is consistently superior to both. 
	
	It is important to point out another methodological finding coming out from this research and that contributes to consolidate an emerging trend. The most suggestive part of Learned Data Structures, i.e., the delegation to  NNs to learn from data,  results to be computationally expensive with respect to the tight loop of Sorted Table Search procedures. Here we use 0-2 layer NNs with ReLU activators.  It is of interest to highlight that an analogous conclusion was reached in regard to Learned Bloom Filters \cite{LearnedBloomEmpirical}, with the use of Recurrent NN. Apparently, the learning and prediction power of the current NNs is a  time performance mismatch with respect to Data Structures that require only a few instructions per iteration in order to answer a query. This asks for NNs and Classifiers that are fast to train and query and effective in learning.

	\section{Binary and Interpolation Search}\label{STSP}
	
	\subsection{Binary Search: Array Layouts and their  Branch-Free Algorithms}\label{sec:BS}
	Following research in \cite{Morin17}, we review here four basic layouts of a table $A$ for Binary Search, as follows.
	
\begin{figure}[tbh]
		\begin{center}
			\includegraphics[width=0.40\textwidth]{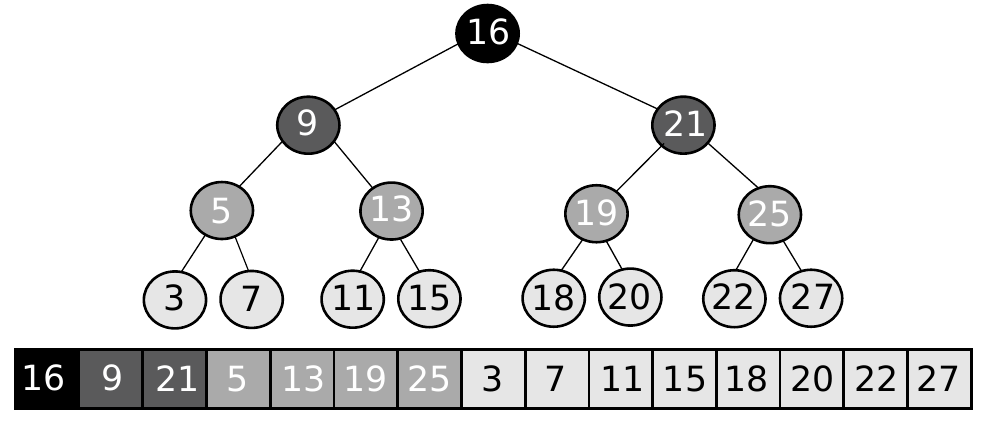}
			\caption{An example of  Eyzinger layout of a table with 15 elements  (see also \cite{Morin17}).}
			\label{fig:EyL}
		\end{center}
	\end{figure}

\begin{figure}[tbh]
		\begin{center}
			\includegraphics[width=0.50\textwidth]{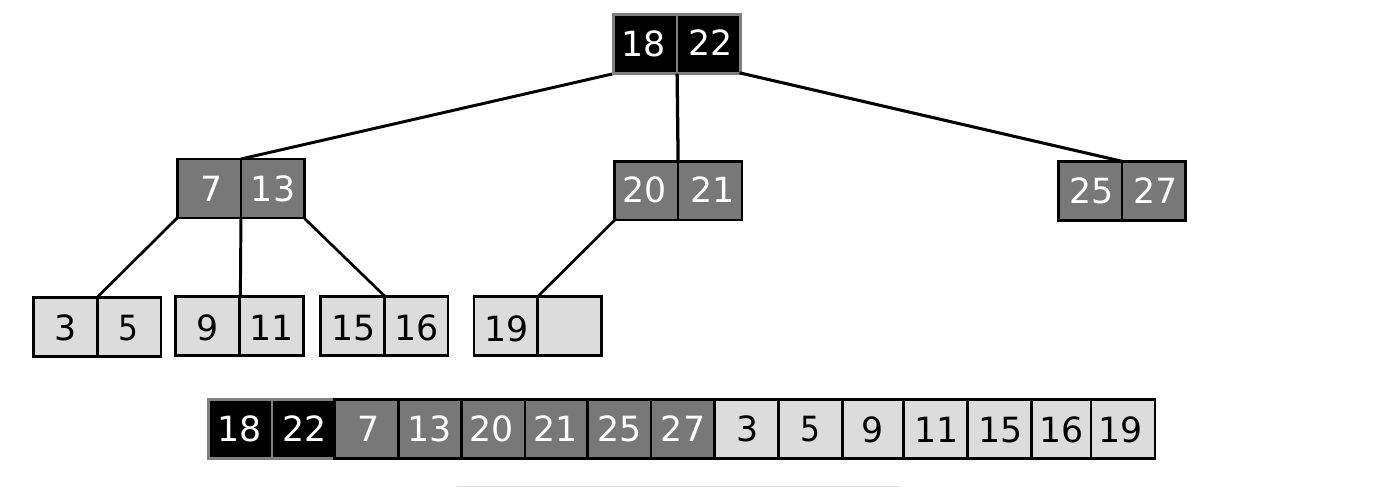}
			\caption{An example of $(B+1)$ layout of a table  with 15 elements and $B=2$ (see also \cite{Morin17}).}
			\label{fig:B-Tree}
		\end{center}
	\end{figure}

\begin{figure}[tbh]
		\begin{center}
			\includegraphics[width=0.40\textwidth]{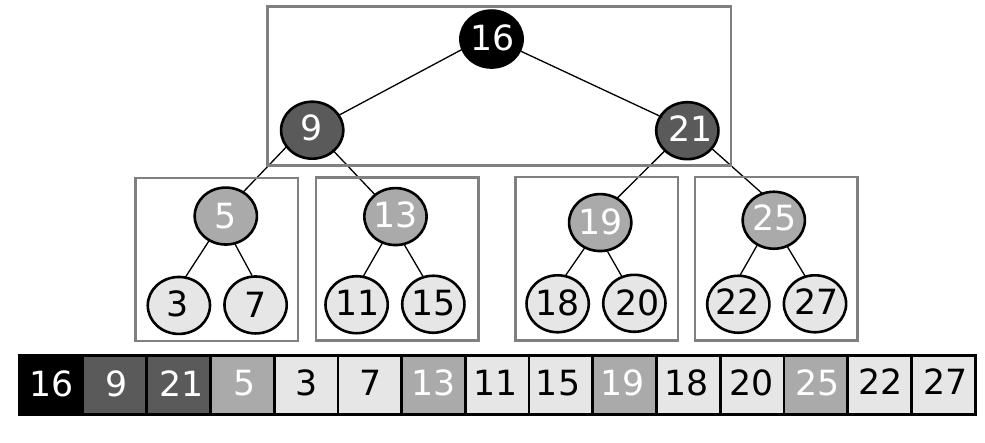}
			\caption{An example of  Van Edme Boas layout of a table with 15 elements (see also \cite{Morin17}).}
			\label{fig:VEB}
		\end{center}
	\end{figure}

	\begin{itemize}
		\item [1] Sorted. It is the classic textbook layout for standard Binary Search. In particular, We consider the branch-free implementation provided in [11], referring to it as BFS, with prefetching (see Algorithm 1). The term branch-free refers to the fact that there is no branching in the while loop. Indeed, as explained in [11], the case statement in line 8 of Algorithm 1 is translated into a conditional move instruction that does not alter the flow of the assembly program corresponding to that C++ code. In turn, that has the advantage of a better use of instruction pipelining with respect to the branchy version of the same code. Prefetching, i.e., instructions 6 and 7 in Algorithm 1, refers to the fact that the procedure loads in the cache data ahead of their possible use. As shown in [11], such an implementation of Binary Search is substantially faster than its branchy counterpart, which we refer to as BBS. 
		
		\item [2] Eytzinger \cite{Morin17}.  The sorted table is now seen as stored in a virtual complete balanced binary search tree. Such a tree is laid-out in Breadth-First Search order in an array. An example is provided in Fig. \ref{fig:EyL}. Also, in this case, we adopt a branch-free version with prefetching of the binary search procedure corresponding to this layout. It is reported in Algorithm \ref{AL:BFE}.  In what follows, we refer to this procedure as {\bf BFE}. 
		
		\item [3] B-tree. The sorted table is now seen as a $B+1$ search tree \cite{comer1979ubiquitous}, which is then laid out in analogy with an Eytzinger layout.  An example is provided in Fig. \ref{fig:B-Tree}. Also, in this case, we adopt a branch-free version of the binary search procedure corresponding to this layout, with prefetching. It is taken from the software associated to \cite{Morin17}. Due to its length, it is not reported here. In what follows, we refer to this procedure as {\bf BFB}. 
		
		\item [4] Van Edme Boas \cite{Prokop}. The sorted table is again seen as stored in a virtual complete binary tree. Such a tree, assuming that it stores $n$ elements and letting $h$ denote its height, is laid out recursively as follows. If $n=1$, the element is stored in $A[0]$.  Else,  the top part of the tree  of height  $\lfloor h/2 \rfloor$ is laid out recursively  in $A[0\cdots,2^{1+\lfloor h/2 \rfloor}-2]$. At the leaves of this top tree, there are at most $2^{1+ \lfloor h/2 \rfloor} $  subtrees that are laid out recursively and from left to right, starting at position $2^{1+\lfloor h/2 \rfloor}-1$ of the array. An example is provided in Fig. \ref{fig:VEB}. It is worth pointing out that the extensive experiments conducted in \cite{Morin17}, as well as further results presented in \cite{MorinWeb}, indicate that such a layout, on occasions and on large datasets,  may be superior to the ones mentioned earlier. Due to such inconsistency in performance and since the intent of this research is methodological,  it will not be included in the experimental part of this study. 
	\end{itemize}
	
	\begin{algorithm}
		\caption{Branch-free implementation of classic Binary Search with prefetching. The code is as in   \cite{Morin17}.}
		\label{AL:BFS}
		\begin{algorithmic}[1]
			\BState int prefetchBranchfreeBS(int *A, int x,  int left, int right)\{
			\State \ \ \ const int *base = A;
			\State \ \ \ int n = right;
			\State \ \ \ while (n $>$ 1) \{
			\State	\ \ \ \ \ const int half = n / 2;
			\State	\ \ \ \ \ \_\_builtin\_prefetch(base + half/2, 0, 0);
			\State	\ \ \ \ \ \_\_builtin\_prefetch(base + half + half/2, 0, 0);
			\State	\ \ \ \ \ base = (base[half] $<$ x) ? \&base[half] : base;
			\State	\ \ \ \ \ n -= half;
			\State	\ \ \ \}
			\State \ \ \ return (*base < x) + base - A;
			\BState \}	
		\end{algorithmic}
	\end{algorithm}
	
	\begin{algorithm}
		
		\caption{Branch-free implementation of Binary Search with Eyzinger layout and prefetching. The code is as in   \cite{Morin17}.}
		\label{AL:BFE}
		\begin{algorithmic}[3]
			\BState int eytzBS(int *A, int x,  int left, int right)\{
			\State \ \ \ int i = 0;
			\State \ \ \ int n = right;
			\State \ \ \ while (i $<$ n)\{
			\State \ \ \ \ \ \_\_builtin\_prefetch(A+(multiplier*i + offset)); 
			\State \ \ \ \ \ i = (x $<=$ A[i]) ? (2*i + 1) : (2*i + 2);
			\State \ \ \ \}
			\State \ \ \ int j = (i+1) $>>$ \_\_builtin\_ffs($\sim$(i+1));
			\State \ \ \ return (j == 0) ? n : j-1;
			\BState \}	
		\end{algorithmic}
	\end{algorithm}

	\subsection{Interpolation Search}
	Such a Sorted Table Search technique was introduced by Peterson \cite{Peterson57} and it has received some attention in terms of analysis \cite{dataStructHand} since it works very well on data following a Uniform or nearly Uniform distribution and very poorly on data following other distributions. The procedure adopted here is reported in 
	Algorithm \ref{AL:IBS}.  In what follows, we refer to such a procedure as {\bf IBS}.  It is of interest to point out that whether a branch-free implementation of this method is advantageous with respect to its branchy version has not been extensively investigated.  Experiments conducted for this study (available upon request)  show that there no advantage. As for prefetching, due to the very irregular access pattern that this procedure has to the table, it is obvious that it brings no advantage.

	For completeness, we mention that modified versions of Interpolation Search have been designed, both in the static and dynamic case (see  \cite{KAPORIS20,Willard85} and references therein). Those versions can be shown to have a   $O(log log n )$ or even an $O(1)$  expected time performance if the data is drawn from suitable distributions that exhibit some mathematically defined notion of regularity (see references for details). Somewhat unfortunately, those variants, although of theoretic interest,  do not seem to be well suited to compete with standard Interpolation Search in real settings. In particular, the (static) variants proposed by Willard depend on three parameters $\alpha$, $\Theta$ and $\Phi$ that are used to regularize the runtime of the procedure and that must be set through an empirical analysis. Another problem is the introduction of some arithmetic and selection operations on floating-point data types that are computationally expensive to calculate on today's CPUs and introduce a substantial overhead in addition to element search operations. 
	More recently, heuristic and somewhat engineered variants of Interpolation Search have been introduced in \cite{VanSandt19}. To the best of our experience with the available code and with our own implementation of the heuristics, they may err when an element searched for is not in the table.  For the reasons just outlined for each of them, the variants mentioned above are not included in this study.

	\begin{algorithm}
		\caption{Implementation of Classic Interpolation Search.}
		\label{AL:IBS}
		\begin{algorithmic}[2]
			\BState int interpolation\_Search(int *arr, int x, int start, int end)\{
			\State \ \ \ int lo = start, hi = (end - 1);
			\State \ \ \ while (lo $<=$ hi \&\& x $>=$ arr[lo] \&\& x $<=$ arr[hi]) \{
			\State	\ \ \ \ \  if (lo == hi) \{
			\State	\ \ \ \ \ \ \ if (arr[lo] == x) return lo;
			\State	\ \ \ \ \ \ \ return -1;
			\State	\ \ \ \ \ \}
			\State	\ \ \ \ \ int pos = lo + (((double)(hi - lo) / (arr[hi] - arr[lo])) * (x - arr[lo]));
			\State	\ \ \ \ \ if (arr[pos] == x) return pos;
			\State	\ \ \ \ \ if (arr[pos] $<$ x) lo = pos + 1;
			\State	\ \ \ \ \ \ \ else hi = pos - 1;
			\State	\ \ \ \}
			\State \ \ \ return pos;
			\BState \}	
		\end{algorithmic}
	\end{algorithm}
	
	\section{Regression Analysis, Learning Functions, and  Sorted Table Search}\label{sec:LST}
	
	It is well known that  Sorted Table  Search can be phrased as  Predecessor Search Problem ({\bf PSP}, for short):  for a given query element $x$, return the index $j$ such that $A[j] \leq x < A[j+1]$. Kraska et al. \cite{kraska2018case} have proposed an approach that transforms {\bf PSP} into a  Regression Analysis  problem ({\bf RA}, for short) \cite{FreedmanStat}.  In what follows, we first review the regression techniques used for this research (Section \ref{sec:regression}) and then cast {\bf PSP} into an  {\bf RA} (Section \ref{sec:PS}).

	\subsection{Regression Analysis}\label{sec:regression}
	
	It  is a methodology for estimating a given  function $F:\mathbb{R}^m \rightarrow \mathbb{R}$  via a specific function model $\tilde{F}$. The independent variables in $x \in \mathbb{R}^m$ and the dependent variable $y \in \mathbb{R}$  are usually referred to as predictors and outcome, respectively. The  parameters of $\tilde{F}$ are estimated  by minimizing an error function,
	computed using a sample set of predictors-outcome measurements.
	The most commonly used Regression Loss Function is the Mean Square Error. Such a task can be accomplished in several ways. Here we follow the methods outlined in  \cite{Goodfellow-et-al-2016}. In particular, we first present closed-form formulae solving the posed minimization problem, with a linear (as a matter of fact, polynomial) model (Section \ref{sec:linear}). Then, we outline a gradient descent method based on NN (Section \ref{sec:NN}).

	\subsubsection{Multivariate  Linear  Regression }\label{sec:linear}
	Linear Regression ({\bf LR}, for short) is a specific approach to the regression that assumes a linear function (i.e. polynomial of degree 1) as a model. The case of one predictor is referred to as Simple Linear Regression ({\bf SLR}, for short), and otherwise as Multivariate Linear Regression ({\bf MLR}, for short).
	
	For  {\bf MLR}, given  a sample set of $n$ predictor-outcome couples $(\mathbf{x}_i,y_i)$, where $\mathbf{x}_i \in \mathbb{R}^m$ and $y_i \in \mathbb{R}$, the goal is to characterize the linear function model $\tilde{F}(\mathbf{x})=\mathbf{\hat{w}} \mathbf{x}^T+\hat{b}$ by estimating the parameters $\mathbf{\hat{w}} \in \mathbb{R}^m$ and $\hat{b} \in \mathbb{R}$, using the sample set. We can define  a matrix $\mathbf{Z}$ of size $n \times (m+1)$ (usually referred to as the design matrix), where $\mathbf{Z}_i$ 
	is the $i$-th row of $\mathbf{Z}$ such that ${\mathbf{Z}_i}= [\mathbf{x}_i,1 ]$. Moreover, $\mathbf{y}$ indicates the vector of size $n$ such that the outcome  $y_j$ is its $j$-th component. The Mean Square Error  minimization on the basis of the estimation is:
	
	\begin{equation}
	{\bf MSE}(\mathbf{w},b) = {\frac{1}{n} \left \| [\mathbf{w},b]  \mathbf{Z}^T-\mathbf{y} \right \|}_2^2
	\label{MSE}
	\end{equation}
	
	{\bf MSE} is a convex quadratic function on $[\mathbf{w},b]$, so that the unique values  that minimize it  can be obtained by setting its gradient $\nabla_{\mathbf{w},b}$ equal  to zero.
	The closed form solution for the parameters $\mathbf{w},b$ is  
	
	\begin{equation}
	[\mathbf{\hat{w}},\hat{b}]= \mathbf{y} \mathbf{Z} (\mathbf{Z}^T \mathbf{Z})^{-1} 
	\label{cfreg}
	\end{equation}
	
	It is to be noted that, 
	for  {\bf SLR},  the case of models more complicated than a polynomial of degree one, specifically polynomial models with degree $g>1$,  can be reduced to {\bf MLR}. Indeed, we can consider the model 
	
	$$\tilde{F}(\mathbf{z})=\sum_{i=1}^g w_i x^i + b=\mathbf{w}\mathbf{z}^T+b, $$
	
	\noindent where $w$ is of size $g$, $\mathbf{z}=[x,..,x^{g-1},x^g] \in \mathbb{R}^{g}$ is the predictor vector for {\bf  MLR}.
	
	\subsubsection{Multi Layer Feed Forward Neural Networks with ReLU Activators  for Regression}\label{sec:NN}
\begin{figure}[t]
		\begin{center}
			\includegraphics[width=3.5cm]{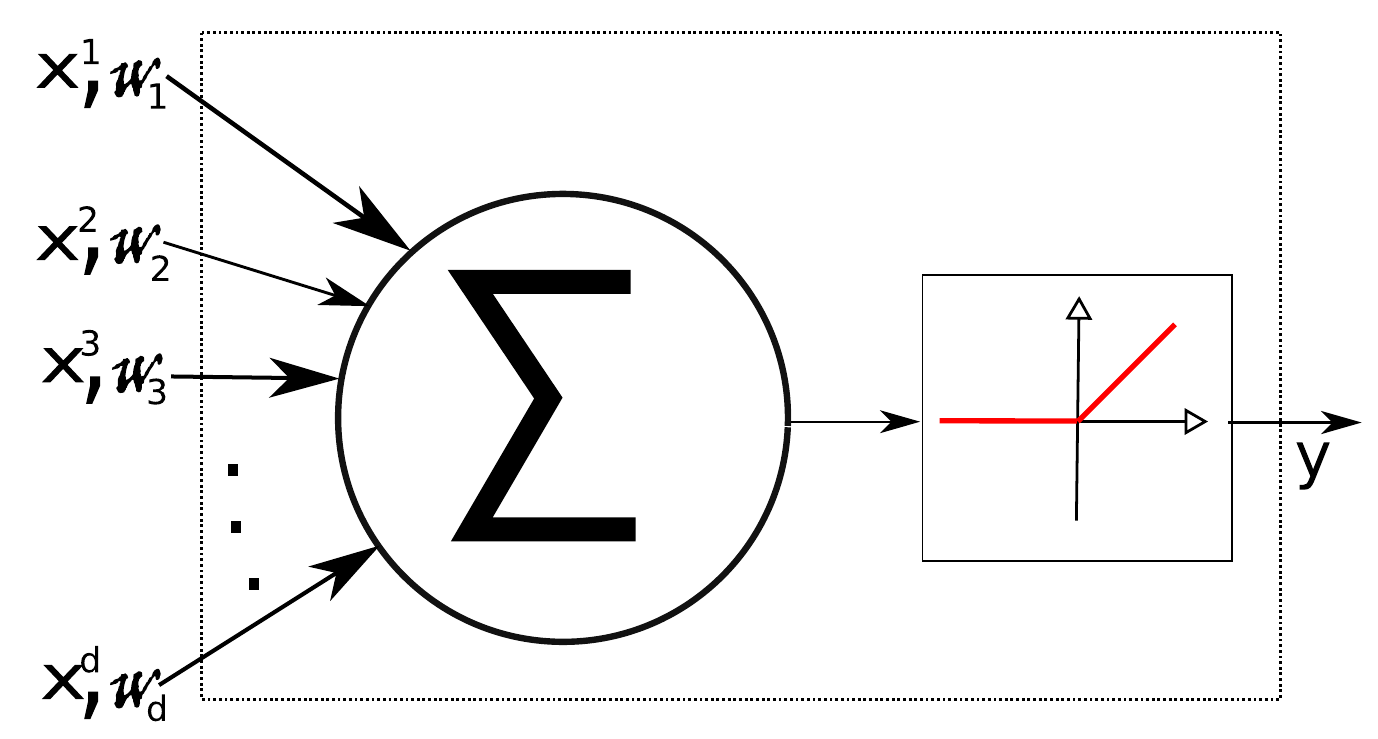}
			\caption{The atomic element uses $(x^1,..,x^d)$ and $ (w_1,..,w_d)$ as inputs. Each $x^i$ is a  binary digit while each $w_i$ is a real value. The final output $y=max(0,\Sigma_{i=1}^d w_i x^i)$.}
			\label{neurone}
		\end{center}
		
	\end{figure}

	Another approach to regression is to use a universal approximator such as a neural network. Here, we concentrate on \emph{feed forward neural networks} to learn a function $F$ from a sample set $(\mathbf{x}_i,y_i)$.
	The general strategy is simple. Indeed, NN uses a \emph{learning phase} which is iterative,  starting from an initial approximation $\tilde{F}_0$. At each step $i$, the current approximation is refined, using the sample set, leading to a new solution $\tilde{F}_i$ such that $E(\tilde{F}_i(\mathbf{x}),y) \le E(\tilde{F}_{i-1}(\mathbf{x}),y)$, where $E$ represents a suitably chosen error function. The process halts when we have only marginal gains on the error. That is,  given a  tolerance $\delta$, the process stops when  $\tilde{F}=\tilde{F}_i$ such that $|E(\tilde{F}_i(\mathbf{x}),y) - E(\tilde{F}_{i-1}(\mathbf{x}),y)| \le \delta$. 
	The approach to regression by an NN involves the representation of the input $\mathbf{x}$ into a binary string of size $d$ of each of its components. The case of interest here is univariate and we denote by $\vv{x}$ such a binary string, outlining the relevant features of the NN next \cite{Bishop95}:

	\begin{itemize} 
		\item[(1)] {\bf ARCHITECTURE TOPOLOGY}. It is characterized by the following:
		\subitem (1.a)  {\bf The neuron, i.e. the atomic element of the {\bf NN}}.  A scheme is provided in Figure \ref{neurone}.  The basic operations on its $2d$ binary input lines and the activation via  function  $f$ are relevant. For the first, we use the scalar product between the input vectors  $\vv{w}$ and  $\vv{x}$. The vector $\vv{w}$ is  referred to as  \emph{weight}. For the second, we use  is the so called \emph{reLU} i.e. $f(x)=max(0,x)$,.
		\subitem (1.b) {\bf The number of hidden layers} $K$.
		\subitem (1.c) {\bf For each layer $i$, its   number $n_i$ of atomic elements}. 
		\subitem (1.d) {\bf The connection between layers}. The layers are fully connected, i.e. each atomic element at layer $K$ is connected to all the atomic elements at layer $K+1$. 
		\item[(2)] {\bf LEARNING ALGORITHM}: This is characterized as follows:
		\subitem(2.a)  {\bf The error function} $E$, used to measure how close the approximation  $\tilde{F}$ is to $F$. 
		
	\end{itemize}
	
	The specific NN models used for our  experiments are:
	
	\begin{itemize}
		
		\item \textbf{NN0}: A zero hidden layer network, i.e.,  an NN composed by only one input layer of $n_0=256$ neurons and an output layer with $1$ neuron.   
		
		\item \textbf{NN1}: A one hidden layer network, i.e.,  one hidden layer composed of $n_1=256$ neurons inserted between the input and output layers of {\bf NN0}. 
		
		\item \textbf{NN2}: A two hidden layer network, i.e., a hidden layer of $n_2=256$ neurons inserted between the input and output layers of NN1. 
		
	\end{itemize}

	\subsection{Predecessor Search as a Regression  Analysis Problem}\label{sec:PS}
	
	Consider a sorted table $A$ of  $n$ keys, taken from a universe $U$. Let $EP$ be the \emph{empirical predecessor}  function  of elements of $U$ with respect to  (sample) $A$. That is, for each $z \in U$,  $EP(z)=|\{y \in A | y \leq z\}|$.
	Notice that knowledge of $EP$ provides a solution to  {\bf PSP}  since, given an element $z \in U$,  only one evaluation of $EP$ provides the predecessor index we are looking for. 
	Usually, $EP$ is not available,  implying that an implicit or explicit mathematical expression that we can use over and over again to solve {\bf PSP} must be \emph{ estimated numerically}.  
	As already mentioned, this is now a Regression Analysis problem, where the sample set is given by the $n$ pairs   $(A[j],j)$. Assuming that an estimate  $\tilde{EP}$ of $EP$ has been found, its precision with respect to the elements in the table $A$ is computed as follows. 
	
\begin{figure}[t]
		\begin{center}
			\includegraphics[width=8.0cm]{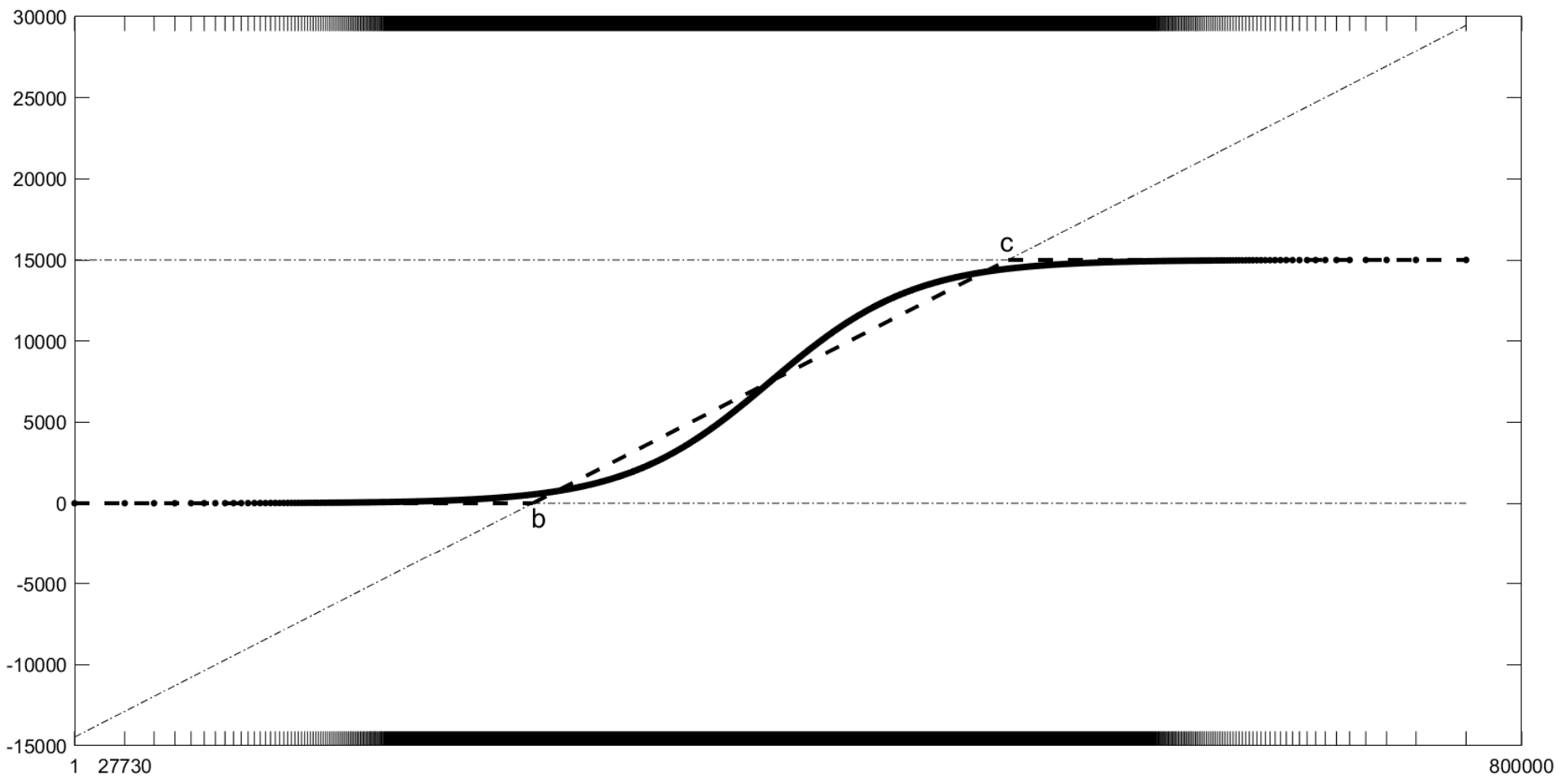}
			\caption{A table $A$ of 15000 elements is shown  on the abscissa (a small vertical line for each element). The curve corresponding to its $EP$ is a sigmoid shown in bold. The two vertical lines delimit the range of indices of the table. The dotten line in bold corresponds to the $\tilde{EP}$   obtained via  {\bf SLR}. Points (b) and (c) denote $\epsilon_2$ and $\epsilon_3$ for that table and prediction. As for $\epsilon_1$, is computed with the use of the regression  line between points (b) and (c)}
			\label{errsigmoide}
		\end{center}
	\end{figure}

	\begin{enumerate}[(a)]
		\item For each $j$ in $[1,n]$  such that $1\leq\tilde{EP}(A[j])\leq n$, let $\epsilon_1=max_j |\tilde{EP}(A[j])-j)|$. Once that  $\tilde{EP}$  and $\epsilon_1$ are  available for $A$, we can solve  {\bf PSP} as follows.  
		For a query $x$, the  value of $\tilde{EP}(x)$  is computed. When $1 \leq \tilde{EP}(x) \leq n $, the search is completed via a standard Sorted Search  Procedure  in the interval $[\tilde{EP}(x)-\epsilon_1, \tilde{EP}(x)+\epsilon_1]$. An example is provided in Fig. \ref{errsigmoide} with an $EP$ that follows a sigmoid function. That is, a function $\sigma(y)=\frac{1}{1+e^{-y}}$.  \label{caseA}
		\item For each $j$ in $[1,n]$ such that $\tilde{EP}(A[j])<1$, let $\epsilon_2$  be the maximum such  a  $j$. Once that  $\tilde{EP}$  and $\epsilon_2$ are  available for $A$, we can  solve  {\bf PSP} as in case (a), when  $\tilde{EP}(x) < 1 $ with the interval $[1, \epsilon_2]$. An example is provided in Fig. \ref{errsigmoide}. 
		\label{caseb}
		\item For each $j$ in $[1,n]$ such that $\tilde{EP}(A[j])>n$, let $\epsilon_3$ be the minimum such a  $j$. Once that  $\tilde{EP}$  and $\epsilon_3$ are  available for $A$, we can  solve  {\bf PSP} as in case (a), when  $\tilde{EP}(x) >n $ with the interval $[\epsilon_3,  n ]$. An example is provided in Fig. \ref{errsigmoide}. 
		\label{casec}
	\end{enumerate}
	
	If we use the technique outlined in Section \ref{sec:linear}, i.e., equation (\ref{cfreg}), it is immediate to see that the estimation of  $\tilde{EP}$ can be performed in polynomial time in $n$, assuming that $m \leq n$.  
	Indeed, $m=1$ for {\bf SLR} and it is a constant for polynomial models of degree at least two.  The approach based on  NNs is more heuristic, in the sense that there is no guarantee to find the optimal solution. Moreover, being an iterative process with no guarantee of convergence, it is quite difficult to provide an estimate of the computational complexity of the entire procedure unless one fixes \emph{a priori} the number of iterations required for halting rather than the achievement of a given level of precision.   
	
	Let now $maxleft= \max{(2\epsilon_1+1, \epsilon_2, \epsilon_3)/n}$. It represents the fraction of the table that is left to search after a prediction is made. In what follows we refer to $1-maxleft$ 
	as the  \emph{reduction factor}. It represents the percentage of the table that the prediction has excluded from further consideration, e.g., the first iteration of binary search has a reduction factor of $50\%$. It is to be noted that, although Sorted  Table Search procedures may benefit in theory from a reduction to be faster than their standard counterparts, in practice there is a trade-off between the cost of prediction and the reduction in size that is attained.
	That is, experimentally, it may as well be that a Learned Sorted Table Search procedure may be slower than its standard counterpart, if the estimate $\tilde{EP}$ is not particularly sharp. In order to study such a trade-off, that provides valuable information for the real use of Learned Sorted Table Search procedures, we construct synthetic datasets (see Section \ref{Error}) and we experiment with them (see Section  \ref{sec:convenient}
	). 
	
	\subsection{Dichotomic Sorted Table Search: Prediction via Regression Analysis Approximations}  
	
	Dichotomic Sorted  Table Search procedures, i.e., the standard and well-known strategies to search in a sorted table,   can also be viewed as procedures that, in order to choose an index,  use an approximation to a prediction that can be made via Regression Analysis.  It is of interest to make explicit such a generic paradigm, which naturally complements the one of the previous section. 

	Consider a black box  $BBM(A,k,l)$ that gets as inputs a sorted table $A$ and two indices $k,l$ of $A$, and returns  a model $M$ for $\tilde{EP}$  computed  on $A[k,l]$. Then, given a query element $x$, $M(x)$ provides an index  $\hat{j}$ in $[l,k]$ that can be used for the \vir{dichotomic decision} in that interval (for clarity of exposition, we are assuming that the prediction always falls within the required interval).    Algorithm  $GDSA$, reported in Figure \ref{AL:GSA},  is the corresponding paradigm to Sorted Table Search via the black box.

	\begin{algorithm}
		\caption{The generic search algorithm, GDSA}
		\label{AL:GSA}
		\begin{algorithmic}[4]
			\BState int {GDSA}(int *arr, int x, int start, int end)\{
			\State \ \ \ int lo = start, hi = (end - 1);
			\State \ \ \ while (lo $<=$ hi \&\& x $>=$ arr[lo] \&\& x $<=$ arr[hi]) \{
			\State	\ \ \ \ \  if (lo == hi) \{
			\State	\ \ \ \ \ \ \ if (arr[lo] == x) return lo;
			\State	\ \ \ \ \ \ \ return (arr[lo]<x ? lo : lo-1);
			\State	\ \ \ \ \ \}
			\State	\ \ \ \ \ M=BBM(arr,lo,hi);
			\State	\ \ \ \ \ pos = M(x);
			\State	\ \ \ \ \ if (arr[pos] == x) return pos;
			\State	\ \ \ \ \ if (arr[pos] $<$ x) lo = pos + 1;
			\State	\ \ \ \ \ \ \ else hi = pos - 1;
			\State	\ \ \ \}
			\State \ \ \ return pos;
			\BState \}	
		\end{algorithmic}
	\end{algorithm}

	Let us consider the case when the $BBM$ returns a model $M$ which uses a \emph{line} of equation $y=\alpha z+\beta$ to make the prediction. In this case, $\hat{j}= M(x) = \left\lceil \alpha x+\beta \right\rceil$. When $M$ is the {\bf SLR}  model  $LRM$ for $EP$, $GDSA$ specializes to a new dicothomic search procedure that, although unlikely to be practical, it is optimal in terms of approximating  the $EP$ function of the data  at each step of the procedure.  The known search procedures are particularly convenient approximations to $LRM$ in terms of Mean Square Error. 
	When $M(x)=\left\lceil \frac{l+k}{2} \right\rceil $, we have a \emph{Binary Search Model} ($BSM$) that specializes $GDSA$ to Binary Search. When $M(x)=F_{g-1}$, where  $g$ is the least integer such that $F_{g+1}\geq (k-l+1)$ and $F_j$  denotes the $j$-th Fibonacci number, we have  a \emph{Fibonacci  Search Model} ($FSM$) that specializes $GDSA$ to Fibonacci  Search (see \cite{KnuthS}).  
	When $$M(x)= \left\lceil k + \frac{l-k}{A[l]-A[k]} (x- A[k]) \right\rceil$$  we have an  \emph{Interpolation Search Model} ($ISM$) that specializes $GDSA$ to Interpolation Search. 
	One can extend the class of models also to \emph{Slope Reuse Interpolation Search} ($SIM$) \cite{VanSandt19}, characterized by the line:
	
	\medskip 
	
	$$M(x)= \left\lceil k + s (x - A[k]) \right\rceil$$
	
	\noindent where $s$ denotes a precomputed slope. 

	Finally, it is worth pointing out that models more complicated  than linear can be used. For instance,  \emph{Three Point Interpolation Search } $TIP$ has a model for $BBM$, which is characterized by  a curve rather than a line \cite{VanSandt19}. The algorithm selects three increasing positions $k \le l \le m$, and found the prediction $\hat{j}$ in the following way:
	
	\begin{multline*}
	M(x)= l+ \\ \frac{(A[l]-x)(l-m)(l-k)(A[m]-A[k])}{(A[m]-x)(l-m)(A[k]-A[l])+A[k](l-k)(A[l]-A[m])}
	\end{multline*}

	\section{Experimental Setup}
	\subsection{Hardware}\label{hardware}
	All the experiments have been performed on a workstation equipped with an Intel Core i7-8700 3.2GHz CPU and an Nvidia Titan V GPU. The total amount of system memory is 32 Gbyte of DDR4. The GPU is also supplied with its own 12 Gbyte of DDR5 memory and adopts a CUDA parallel computing platform and an API. The single computational element of a GPU, referred to as CUDA core, is specifically designed for one single-precision multiply-accumulate operation. It is the basic step of each computational cell of a feed-forward neural network. Therefore, such an architecture very well fits the  NN models chosen for this study \cite{Raina2009}.   
	
	For reference in the following, it is useful to discuss the overhead in data I/O between CPU and GPU. The slowest link involved in GPU computing is memory transfers from system to GPU.
	This is actually a technological limit and it is due to the difference in bandwidths between the system memory, the GPU device memory and the PCI express bus connecting them. In particular, the system memory is limited to the DDR4 generation, which allows at most a bandwidth of $~25$ Gbyte/sec. The PCIe $3$, which is actually the standard, supports a maximum bandwidth of $32$ GB/sec. Such values are undersized with respect to the DDR5 memory bandwidth, which in the case of our GPU device is $~600$ GByte/sec. We expect that such differences will be smoothed out by the introduction of DDR5 system memory and PCI-Express $5$ bus technology. 
	
	\subsection{Datasets}\label{sec:Datasets}
	We use two different types of data. The first, described in Section \ref{dataset1}, can be thought of as a standard in this area of research, e.g., \cite{kraska2018case}. 
	As anticipated in Section \ref{sec:PS}, the second dataset type is designed to get a quantitative estimate of the  reduction factor  that the procedures in that section must attain in order for 
	Learned Sorted Search procedures to be advantageous with respect to their standard counterparts.  Those datasets are described in Section  \ref{Error}. 
	
	\subsubsection{Preliminary Definitions}\label{sec:distri}
	
	It is useful to recall the definitions of Uniform, Log-normal and Logit distributions and the type of $EP$ functions, in terms of CDF,  that correspond to them and which must be eventually learned when tables are built by sampling from those distributions.
	The Uniform distribution $U$ is 
	
	$$U(x,a,b) =
	\left\{
	\begin{array}{ll}
	\frac{1}{b-a}  & \mbox{if } x \in [ a,b ] \\
	0 & otherwise
	\end{array}
	\right.$$
	
	\noindent where $a$ and $b$ are the parameters of the distribution.  The CDF  corresponding to it is a straight line in the interval $[a,b]$. 
	
	The Log-normal distribution $L$ has two parameters: its mean  $\mu$ and standard deviation $\sigma$. It is defined for $x>0$ as follows:
	
	$$L(x,\mu,\sigma)=\frac{e^{-\frac{(lnx-\mu)^2}{2\sigma^2}}}{x\sqrt{2\pi}\sigma}$$
	
	For this study, the parameters are chosen so that the CDF is a sigmoid function, where the concave part is predominant.  This case is representative of  
	other well-known distributions such as Poisson and $\chi^2$. 
	
	The Logit distribution is defined as 
	
	$$G(x,\mu,s)=\frac{e^{-\frac{(x-\mu)}{s}}}{s(1+{e^-{\frac{(x-\mu)}{s}}})}$$
	
	\noindent with mean $\mu$ and standard deviation  $s>0$. Its CDF  corresponds to a sigmoid function. This case is representative of  
	other well-known distributions such as Normal, Binomial and Hypergeometric. 
	
	\medskip 
	
	When sampling from those distributions, we take integers in the interval is $[1, 2^{r-1}-1]$, where $r$ is determined by the precision used to represent integers, i.e., 32 or 64 bits.
	\subsubsection{Dataset1:  Literature Benchmarks}\label{dataset1}
	We use synthetic and real datasets. The first  ones are generated according to the Uniform (with parameters  $a=1$ and $b=2^{r-1}-1$  and Log-normal (with standard parameters  $\mu=0$ and $\sigma=1$) distributions. The second ones are domain-dependent because they are obtained from different application areas (such as IoT or the Web applications). In particular, the real-weblogs \cite{Kraska:FITing, kraska2018case} contains timestamps of about 715M requests performed by a web server during 2016.
	The real-iot dataset \cite{Kraska:FITing, kraska2018case} consists of timestamps of about 26M events recorded during 2017 by IoT sensors deployed in academic buildings. All the datasets are sorted and do not contain duplicate values.
	Their characteristics in terms of size are summarized in  Table \ref{tab:generalDatasetCharacteristics}. 
	The method used to extract query datasets takes as input a dataset U among those described above, and it returns three query datasets of the following size with respect to U: $10\%, 50\%, 80\%$, respectively. Each of those consists of roughly 50\% of items in U.  The notation regarding those query datasets is as follows: $<$dataset name$>$-query$<$percentage$>$, e.g., uni-01-query10  refers to the first uniform dataset in which the size of the query dataset is $10\%$  of it.  For this research, query datasets are not sorted. 
	
	\begin{table}[tbh]
		\centering
		\caption{A summary of the Dataset1.  For each dataset in the collection,  we show: the name we have used (column \textbf{Name}), its size in Kilobyte (column \textbf{Size (KB)}), the number of elements  in it (column  \textbf{Items}), and the type of its elements (final column).  (\textbf{Type}).\label{tab:generalDatasetCharacteristics}}
		\begin{tabular}{|l|l|l|l|}
			\hline
			\multicolumn{4}{|c|}{\textbf{Datasets: Uniform Distribution}}          \\ \hline \hline
			\textbf{Name}   & \textbf{Size (KB)} & \textbf{Items} & \textbf{Type} \\ \hline
			uni-01       &   \num{5.37e+0}            &   \num{5.12E+02}&    integer          \\ \hline
			uni-02       &    \num{8.58E+01}         & \num{8.19E+03}           &   integer            \\ \hline
			uni-03      &    \num{1.10E+04}         &  \num{1.05E+06}    & integer              \\ \hline
			uni-04      &    \num{2.81E+06}         &  \num{2.68E+08}            &integer               \\ \hline \hline
			\multicolumn{4}{|c|}{\textbf{Datasets: Log-normal Distribution}}        \\ \hline \hline
			\textbf{Name}   & \textbf{Size (KB)} & \textbf{Items} & \textbf{Type} \\ \hline 
			logn-01       &     \num{5.13e+0}          & \num{5.12E+02} &integer\\ \hline
			logn-02       &    \num{8.23E+01}   & \num{8.19E+03} &integer\\ \hline
			logn-03      &  \num{1.05E+04}  &  \num{1.05E+06}  &integer\\ \hline
			logn-04      & \num{2.71E+06}  &  \num{2.68E+08}  &integer\\ \hline \hline
			\multicolumn{4}{|c|}{\textbf{Datasets: Real Distribution}}             \\ \hline \hline
			\textbf{Name}   & \textbf{Size (KB)} & \textbf{Items} & \textbf{Type} \\ \hline
			real-wl         &\num{3.48E+05}&\num{3.16E+07}& integer\\ \hline
			real-iot             &  \num{1.67E+05} &\num{1.52E+07}&integer\\ \hline
		\end{tabular}
	\end{table}
	
	\subsection{Datasets2: Reduction Factor Simulations }\label{Error}
	In order to estimate the effectiveness of the Learned Sorted Table Search procedures as a function of the reduction factor, we generate synthetic datasets that may be thought of as being challenging for them to handle.   Binary Search has a strategy that is insensitive to the distribution underlying the dataset, e.g., Uniform or Logit, and the worst-case consists of queries for elements that are not in the table.  We use the Log-Normal distribution. As for Interpolation Search,  its performance depends heavily on the distribution characterizing a given table.  Therefore, we generate datasets according to the three distributions defined in Section \ref{sec:distri} since their CDFs are representative of many common distributions.  For completeness, 
	it is worth mentioning that both the Log-normal and the  Logit distributions are not regular according to the definitions provided in \cite{KAPORIS20,Willard85} and that no lower bound is available on the performance of Interpolation Search on tables extracted from those distributions. However,  their non-regularity can be taken as a strong indication that they are challenging for 
	Interpolation Search.  Details regarding the generation of the tables are provided  Section \ref{sec:Taabler}.  

	As for Binary Search and in regard to the generation of query datasets, we proceed as follows. Given a table $T$, we compute a query dataset $Q$ consisting of one million elements not in $T$. For each $x$  in $Q$, we also generate an artificial table interval in which to search for $x$.  The interval contains the predecessor index of $x$. The length of the interval, equal within each query dataset, allows us to control time performance as a function of predetermined reduction factor values.   As for interpolation Search, we use the same query datasets, ignoring the artificial intervals since the reduction factor is generated via the techniques in Section \ref{sec:regression}.  The details are provided in Section \ref{sec:Taabler}. 
	
	\begin{algorithm}
		\caption{Query Generation for a given Table }\label{A:QG}
		\begin{algorithmic}[1]
			\State //Input($T$, $p$, $m$) $\rightarrow$ $Q$
			\State For $i = 1$ to $m$ do:
			\State \ \ \ \ Let $q_i$ be an element chosen uniformly and at random in $[1,  2^{r-1}-1]$  so that $q_i \notin T$
			\State \ \ \ \ Search for the predecessor index $j$ of $q_i$ in $T$.
			\State \ \ \ \ Choose uniformly and at random an integer $k$ in $[0, \lceil (1-p)*n \rceil]$
			\State \ \ \ \ Let $I$ the table interval obtained by juxtaping $[0, \lceil (1-p)*n \rceil]$ to the $T$ so that $k$ and $j$ coincide
			\State \ \ \ \ Report in $Q$ the extremities of interval $I$ and the element $q_i$.
			\State Return $Q$
			
		\end{algorithmic}
	\end{algorithm}

	\subsubsection{Generation of Tables} \label{SS:DGFT}
	We describe the process for the Log-normal distribution, being the same for the Uniform and Logit. The  parameters of the Uniform and Log-normal distributions are as in Section \ref{dataset1}. As for the  Logit, we use   $\mu=0.5$ and $\sigma=0.04$, since those parameters ensure that the $EP$ of the sampled  data is a sigmoid function. Let $n$ be the number of elements a table must have. Then, that many   integers  are chosen according to  the Log-normal   distribution.  The number \textit{n} of elements is chosen according to the  capacity of each component of the internal memory hierarchy.  Indeed, as pointed out in 
	\cite{Morin17}, such a hierarchy has a subtle and deep impact on the performance of Binary Search. For the computer architecture we are using,  the details of the tables we generate are as follows:
	\begin{enumerate}
		\item [$\bullet$] fitting in L1 cache and GPU Memory: cache size 32Kb and each element represented with 32 bits. Therefore, we choose \textit{n} = 7.5K  (this table is denoted  $L1$).
		\item [$\bullet$] fitting in L2 cache and GPU Memory: cache size  256Kb and each element represented with 32 bits. Therefore, we choose \textit{n} = 63K (this table is denoted  $L2$).
		\item [$\bullet$] fitting in L3 cache and GPU Memory: cache size  8Mb and each element represented with 32 bits. Therefore, we choose \textit{n} = 1.5M (this table is denoted  $L3$).
		\item [$\bullet$] fitting in GPU Memory and PC Main Memory: GPU memory 12Gb and each element represented with 64 bits. Therefore, we choose \textit{n} = 1.25G (this table is denoted $L4$).
		\item [$\bullet$] fitting Workstation Main Memory:  Main memory 32Gb and each element represented with 64 bits. Therefore, we choose \textit{n} = 3.75G (this table is denoted $L5$).
		\vspace{6pt}
	\end{enumerate}
	
	\subsubsection{Query dataset generation for a given table T and reduction factor $p$}\label{sec:Taabler}
	Given a previously created table $T$ and a reduction factor $p$, we create a query dataset of $m$ elements, in which each is not in $T$. We also provide an artificial prediction interval for each of those elements. The details are in Algorithm \ref{A:QG}. It takes as input a table $T$, a reduction factor $p$ and the number of query elements $m$ to generate. It returns a query dataset $Q$, containing the queries elements and the fictitious prediction interval.  Such a procedure is systematically used in Section \ref{sec:convenient} in order to find the breakeven reduction factor point in which Learned Binary Search becomes superior to its standard counterpart. 
	
	\section{Function Model Complexities That Binary and Interpolation Search Can Profitably Use as \vir{Oracles}}\label{sec:exp1}
	With reference to Section \ref{sec:regression}, the models that are considered for this study can be divided into two complexity categories: linear, i.e., {\bf NN0} and {\bf SLR}, and non-linear, i.e., {\bf NN1},  {\bf NN2}  and {\bf MLR} with degree higher than one.
	Due to the simplicity and compactness in terms of code of the search procedures we are considering, it is not clear which model is best suited to be \vir{an oracle} for the mentioned procedures. The aim of the experiments presented in this Section is to shed light on this aspect with the use of benchmark datasets in this area, i.e.,  the ones described in Section \ref{dataset1}. 
	
	\subsection{The Cost of Learning: Multivariate Linear Regression  is Better Than NNs}
	Models need to learn the function one is trying to approximate. For the uses intended in this research, two indicators are important. The time required for learning and the reduction factor that one obtains. Indeed, it is worth recalling from Section \ref{sec:PS} that, once an index has been predicted,  a percentage of the table corresponding to its reduction factor is no longer considered for searching. As a comparison baseline for training time, we adopt the time that it takes to sort the table being learned since such a step can intuitively be assumed to be the \vir{learning step} of a Sorted Table  Search procedure.   
	
	For the  training of NNs, we have used {\bf Tensorflow} \cite{tensorflow} with GPU support, while for {\bf SLR} and {\bf MLR}, we use our own  C++ implementation of the procedure outlined in Section \ref{sec:linear}. The sorting routine is the standard C++ {\bf Qsort} utility \cite{Qsort}.  A representative synopsis of the entire set of experiments  is  reported in Tables \ref{T:NNL} and \ref{T:RL}. 
	Based on the performed experiments,  it is clear that  NNs are not competitive. They are slower to train with respect to {\bf SLR}  and {\bf MLR}  models and do not seem to be better in terms of approximation, even if they use a highly engineered platform with GPU support.  Without the use of a GPU, the time performance of their training would considerably worsen (experiments not shown and available upon request). However, GPU has a limited amount of memory, which makes it unusable on large datasets (experimental data not shown and available upon request). 
	On the other hand, {\bf SLR} and {\bf MLR}  models have a  training time comparable with the pre-processing step of sorting the table, yielding no substantial overhead to such a step. 
	This aspect is very important for Decision Support in Main Memory applications. Indeed, as indicated in \cite{rao1999cache}, in that important Data Bases domain one can use sorted tables and binary search as long as the table can be quickly rebuilt to accommodate for changes.  In our setting, that would involve a new training phase.
	
	\begin{table}
		\centering
		\caption{NN  training  with the use of {\bf Tensorflow} on GPU. The first column indicates the dataset, the second the time per elements in seconds, while the  third gives the table reduction expressed in percentage. For timing comparisons, the standard C++ {\bf Qsort} utility, used to sort the mentioned  datasets,  takes $\approx \num{1,19E-07}$s per item.{\large {\tiny }} }\label{T:NNL}
		\begin{tabular}{|c|l|c|}
			\hline
			\multicolumn{3}{|c|}{\textbf{NN0}} \\ \hline \hline
			\multicolumn{1}{|c|}{\textbf{Dataset}}  &
			\multicolumn{1}{c|}{\textbf{Training Time (s)}} & \multicolumn{1}{c|}{\textbf{\% Reduction Factor}} \\ \hline
			uni03 & \num{2,55E-04} & \num{94.08}  \\ \hline
			logn03  & \num{1,39E-04} & \num{54.40}  \\ \hline
			real-wl & \num{2,50E-04} & \num{99.99}  \\ \hline
			real-iot  & \num{1,28E-04} & \num{89.90} \\ \hline
			\multicolumn{3}{|c|}{\textbf{NN1}} \\ \hline \hline
			\multicolumn{1}{|c|}{\textbf{Dataset}}  &
			\multicolumn{1}{c|}{\textbf{Training Time (s)}} & \multicolumn{1}{c|}{\textbf{\% Reduction Factor}} \\ \hline
			uni03  & \num{4,18E-04} & \num{99.89}  \\ \hline
			logn03 & \num{3,79E-04} & \num{94.21}  \\ \hline
			real-wl  & \num{2,31E-04} & \num{99.88}  \\ \hline
			real-iot   & \num{4,20E-04} & \num{98.54}  \\ \hline
			\multicolumn{3}{|c|}{\textbf{NN2}} \\ \hline \hline
			\multicolumn{1}{|c|}{\textbf{Dataset}}  &
			\multicolumn{1}{c|}{\textbf{Training Time (s)}} & \multicolumn{1}{c|}{\textbf{\% Reduction Factor}} \\ \hline
			uni03  & \num{4,49E-04} & \num{99.87}  \\ \hline
			logn03 & \num{8,60E-04} & \num{97.14}  \\ \hline
			real-wl & \num{2,33E-04} & \num{99.84}  \\ \hline
			real-iot  & \num{3,57E-04} & \num{97.31}  \\ \hline	
		\end{tabular}
	\end{table}
	
	\begin{table}
		\centering
		\caption{{\bf SLR} and {\bf MLR} Models. The Table Legend is as in Table \ref{T:NNL}.}\label{T:RL}
		\begin{tabular}{|c|l|c|}
			\hline
			\multicolumn{3}{|c|}{\textbf{SLR}} \\ \hline \hline
			\multicolumn{1}{|c|}{\textbf{Dataset}}  &
			\multicolumn{1}{c|}{\textbf{Training Time (s)}} & \multicolumn{1}{c|}{\textbf{\% Reduction Factor}} \\ \hline
			uni03  & \num{8,20E-08} & \num{99.94}  \\ \hline
			logn03 & \num{5,61E-08} & \num{77.10}  \\ \hline
			real-wl  & \num{5,82E-08} & \num{99.99}  \\ \hline
			real-iot   & \num{7,70E-08} & \num{96.48}  \\ \hline
			\multicolumn{3}{|c|}{\textbf{Quadratic MLR}} \\ \hline \hline
			\multicolumn{1}{|c|}{\textbf{Dataset}}  &
			\multicolumn{1}{c|}{\textbf{Training Time (s)}} & \multicolumn{1}{c|}{\textbf{\% Reduction Factor}} \\ \hline
			uni03  & \num{1.27e-07} & \num{99.98}  \\ \hline
			logn03 & \num{1.02e-07} & \num{90.69}  \\ \hline
			real-wl & \num{1.14e-07} & \num{99.99}  \\ \hline
			real-iot  & \num{1.25e-07} & \num{99.10}  \\ \hline
			\multicolumn{3}{|c|}{\textbf{Cubic MLR}} \\ \hline \hline
			\multicolumn{1}{|c|}{\textbf{Dataset}}  &
			\multicolumn{1}{c|}{\textbf{Training Time (s)}} & \multicolumn{1}{c|}{\textbf{\% Reduction Factor}} \\ \hline
			uni03  & \num{1.84e-07} & \num{99.97}  \\ \hline
			logn03 & \num{1.74e-07} & \num{95.76}  \\ \hline
			real-wl & \num{1.24e-07} & \num{99.45}  \\ \hline
			real-iot  & \num{1.63e-07} & \num{98.87}  \\ \hline
			
		\end{tabular}
	\end{table}
	
	\begin{table}
		\centering
		\caption{Prediction Effectiveness Baseline. The Table Legend follows the terminology introduced in the main text. The timing is reported as time per query in seconds. }\label{T:PB1}
		\begin{tabular}{|c|l|l|l|l|}
			\hline
			\multicolumn{1}{|c|}{\textbf{Dataset-Query}}  &
			\multicolumn{1}{|c|}{\textbf{BFE}}  &
			\multicolumn{1}{c|}{\textbf{L-BFS}} \\ \hline
			uni03 50\%  & \num{1,10E-07} & \num{9,42E-08} \\ \hline
			logn03 50\%  & \num{1,08E-07} & \num{1,60E-07} \\ \hline
			real-wl 50\%  & \num{1,79E-07} & \num{5,05E-08} \\ \hline
			real-iot 50\%  & \num{1,64E-07} & \num{8,32E-08}\\ \hline
		\end{tabular}
	\end{table}
	
	\begin{table}
		\centering
		\caption{Prediction Effectiveness for Baseline. The Table Legend follows the terminology introduced in the main text. The timing is reported as time per query in seconds. }\label{T:PB2}
		\begin{tabular}{|c|l|l|l|l|}
			\hline
			\multicolumn{1}{|c|}{\textbf{Dataset-Query}}  &
			\multicolumn{1}{c|}{\textbf{IBS}} &
			\multicolumn{1}{c|}{\textbf{L-IBS}} \\ \hline
			uni03 50\%  &  \num{7,71E-08} & \num{1,67E-08}  \\ \hline
			logn03 50\%  &  \num{3,87E-07} & \num{9,38E-08}  \\ \hline
			real-wl 50\%  &  \num{9,42E-08} & \num{1,70E-08}  \\ \hline
			real-iot 50\%  & \num{8,68E-07} & \num{2,51E-08} \\ \hline
		\end{tabular}
	\end{table}
	
	\begin{table}
		\centering
		\caption{Prediction Effectiveness-Neural Networks Models. \textbf{NN0-BFS}  refers to Binary Search with 
			{\bf NN0} as an \vir{oracle}, while the other two columns refer to the time taken by {\bf NN1}  and {\bf NN2} to predict the search interval only. The remaining part of the Table Legend is as in Table  \ref{T:PB1}. When the model and the queries are too big to fit in main memory, we report a space error (CPU or GPU)}\label{T:NNP}
		\begin{tabular}{|c|l|l|l|}
			\hline
			\multicolumn{1}{|c|}{\textbf{Dataset-Query}}  &
			\multicolumn{1}{|c|}{\textbf{NN0-BFS}}  &
			\multicolumn{1}{c|}{\textbf{NN1}} & \multicolumn{1}{c|}{\textbf{NN2}} \\\hline
			uni03 50\%  & \num{1,31E-07} & \num{1.56e-06} & \num{5.16e-06}  \\ \hline
			logn03 50\%  & \num{1,92E-07} & \num{1.69e-06} & \num{5.24e-06}   \\ \hline
			real-wl 50\%  & \num{4,59E-07} & Space Error & Space Error   \\ \hline
			real-iot 50\%  & \num{4,76E-07} & \num{1.90e-06} & \num{1.94e-05}   \\ \hline
		\end{tabular}
	\end{table}
	
	\subsection{Simple Linear Regression is the \vir{Oracle} of Choice}
	\subsubsection{Synopsis}
	Once that we have \vir{an oracle} whose prediction, given a query, confines the search to a restricted part of the table, it is important to assess how useful such an approach is in speeding up Binary and Interpolation Search. To this end, we take as baseline {\bf BFE}, which, as indicated in \cite{Morin17}, seems to be the best method with the used dataset sizes and {\bf IBS}. 
	
	As for the \vir{search with an oracle}, 
	we take as baseline  {\bf SLR}  prediction (denoted simply {\bf L} )  with {\bf BFS} and {\bf IBS}.
	We also consider other models as \vir{oracles}. In particular, the ones that have been discussed in the previous section. 
	
	We have performed experiments on all of the datasets and query sets described in Section \ref{dataset1}. We report only a representative synopsis of them, 
	in Tables \ref{T:PB1},\ref{T:PB2}, \ref{T:NNP} and \ref{T:RP}. They refer to queries containing a number of elements equal to 50\% of the reference dataset dimension.
	Coherently with the full set of experiments, the synopsis shows that only {\bf SLR} is competitive as an \vir{oracle} to improve Binary and Interpolation Search. Therefore, from now on, we concentrate only on {\bf SLR}.
	
	\begin{table}
		\centering
		\caption{Prediction Effectiveness-Polynomial Regression Models. The Table reports result with Quadratic (denoted {\bf Q}) and Cubic (denoted {\bf C}) regression. The timing results for {\bf IBS} are analogous and omitted for brevity. The Table Legend is as in Table  \ref{T:PB1}. }\label{T:RP}
		\begin{tabular}{|c|l|l|}
			\hline
			\multicolumn{1}{|c|}{\textbf{Dataset-Query}}  &
			\multicolumn{1}{c|}{\textbf{Q-BFS}} &
			\multicolumn{1}{c|}{\textbf{C-BFS}} \\ \hline
			uni03 50\% & \num{8,11E-08} & \num{9,39E-08} \\\hline
			logn03 50\% & \num{1,59E-07} & \num{1,54E-07} \\\hline
			real-wl 50\% & \num{2.12e-7} & \num{1.80e-7} \\\hline
			real-iot 50\% & \num{1.99e-7} & \num{2.57e-7} \\\hline
		\end{tabular}
	\end{table}
	
	\begin{figure}[ht]
		\begin{center}		\includegraphics[width=0.45\textwidth]{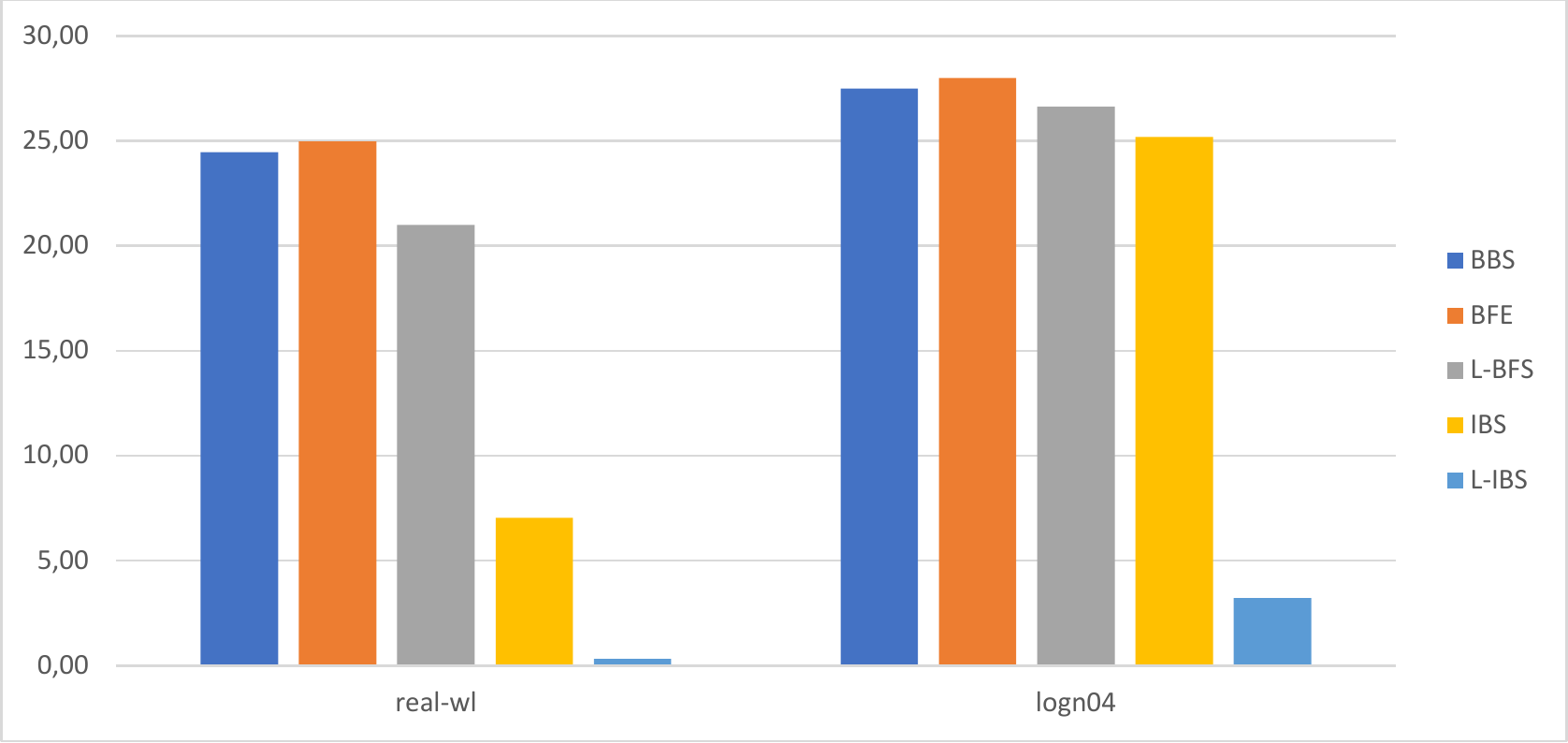}
			\caption{Average per query of the number of iterations for each method on two datasets (y axis). In both cases, the learned procedures perform less iterations than their standard counterparts. Yet, in the second case, {\bf L-BFS} is worse than  {\bf BFE}.  {\bf BBS}  indicates the standard Binary Search procedure.}
			\label{Fig:IT1}
		\end{center}
	\end{figure}

	\subsubsection{A Detailed Analysis of    Simple Linear Regression as an 	\vir{Oracle}}\label{SSS:Oracle}
	In order to synthetically show the full set of experimental results on {\bf Dataset1}, we use the following criterion. Each experiment is labelled with TRUE when the learned routine is better than the standard one, else with FALSE. Finally, in Table \ref{T:MC}, we report only the FALSE cases.   Interestingly, we have that {\bf L-IBS} is always better than its standard counterpart, while there are instances in which {\bf L-BFS} loses to its standard counterpart. In order to get an insight into such a behaviour, we have conducted experiments to monitor the savings in terms of the number of iterations achieved by the \vir{learned} procedures with respect to their standard counterparts. Although a reduction in the interval size to search in yields a gain in terms of iterations, as elementary theory indicates, such a gain is not enough to offset the additional time spent by the earned procedures to make the prediction. An example is illustrated in  Figure \ref{Fig:IT1}, where we  have taken a dataset in which the learned procedures  \vir{win}  and another in which 
	{\bf L-BFS} loses. For completeness, we also include Interpolation Search.  In both cases, the query dataset is $50\%$ of the entire table. 
	
	\begin{table}
		\centering
		\caption{Results of  timing performance on the Benchmark Datasets of Section  \ref{dataset1}. Only {\bf BFS} FALSE results are reported. All results for  {\bf IBS}  return TRUE.}\label{T:MC} 
		\begin{tabular}{|c|l|}
			\hline
			\multicolumn{1}{|c|}{\textbf{Dataset}}  &
			\multicolumn{1}{c|}{\textbf{C1}}  \\ \hline
			uni04 50\%  & FALSE  \\ \hline
			uni04 80\%  & FALSE  \\ \hline
			logn01 10\%  & FALSE  \\ \hline
			logn03 10\%  & FALSE \\ \hline
			logn03 50\%  & FALSE \\ \hline
			logn03 80\%  & FALSE\\ \hline
			logn04 10\%  & FALSE\\ \hline
			logn04 50\%  & FALSE \\ \hline
			logn04 80\%  & FALSE\\ \hline
		\end{tabular}
	\end{table}
	
	\section{A Deeper Look at the Convenience of Learned Sorted Table Search Procedures} \label{sec:convenient}
	The results of the previous section are certainly encouraging, yet they leave unresolved the issue of how small a reduction factor one can afford in order to have learned procedures that are superior to the standard ones. In regard to Binary Search, due to its data-independent nature, it is possible to devise a simulation in which we search the breakeven reduction factor point, i.e., the minimum reduction factor for which Learned Binary Search is better than its standard counterpart, up to $99.95\%$. To this end, by varying the reduction factor, we generate datasets as discussed in Section \ref{Error}, sampling from the Log-normal distribution since for other distributions the results are the same.  In addition to  {\bf BFE} and {\bf BBS}, we have also used the procedure with a B-tree layout (see Section \ref{sec:BS}).  We have also included the time it takes for a simple parallel GPU implementation of Binary Search to process a batch of queries. The motivation is as follows. In their seminal paper  \cite{kraska2018case}, Kraska \emph{et. al.} discuss the potential advantages of GPU use in the realm of Learned Data Structures, in particular for training NN models. Yet, if a GPU is available, it can also be used for Binary Search. Therefore, it is of interest to establish if Learned Sorted Table Search procedures in CPU can outperform a simple  GPU implementation of Binary Search. On the other hand, it is of interest to establish whether GPU implementations of Learned Binary Search can be profitable.  The results are reported in Table \ref{T:BFakeLearn}, with the exception of the B-tree layout since it is inferior to both {\bf BFE} and  {\bf BBS}. The results of the GPU implementation of Learned Binary Search is not reported since its time performance is quite close to the simple GPU  implementation. This is due to the fact that the computation is I/O-bound (CPU-GPU data transmission- see discussion in Section \ref{hardware}).
	
	Interestingly, for datasets that exceed the cache hierarchy capacity, i.e, datasets L4 and L5, the Learned Binary Search procedure is unlikely to be of any use unless the table is really easy to learn via {\bf SLR}. The situation slightly improves for datasets fitting into the cache. Unfortunately, when a GPU is available, the simple-minded GPU implementation of Binary Search is the method of choice. That is, if the hardware is available for training, its availability must be considered also for parallel implementations of the algorithms at hand and accounted for in assessing to what extend Learned Data Structures are of value. In the case of Binary Search, the answer is negative. 

	\begin{table}
		\centering
		\caption{An estimate of how learnable a table must be for { \bf L-BFS}  to be of advantage. For each dataset, the column legend is as follows. The first column  reports the time  for {\bf BFE} or {\bf BBS}, where an asterisk denotes when the latter is better than the former. The second column report the breakeven reduction factor(BRF)  above which {\bf L-BFS} is superior to {\bf BFE} or {\bf BBS}. The third column reports  the time for {\bf L-BFS} at the breakeven point. The final column reports the time of a simple implementation of Binary Search on GPU. All times are per query in seconds.  A dash  indicates that time is not available.  	
		}\label{T:BFakeLearn}
		\begin{tabular}{|c|l|l|l|l|}
			
			\hline
			\multicolumn{5}{|c|}{\textbf{Log-normal}} \\ \hline \hline
			&
			\multicolumn{1}{|c|}{\textbf{BFE/BBS*}}  &
			\multicolumn{2}{c|}{\textbf{L-BFS}} &
			\multicolumn{1}{c|}{\textbf{GPU}} \\ \hline
			& Time (s) & \% BRF & Time (s) & Time (s) \\ \hline
			L1  & \num{8,46E-08} & \num{98.35} & \num{8.31e-8} & \num{3,17E-09} \\ \hline
			L2 & \num{1,08E-07} & \num{99.25}  & \num{1.07e-7} & \num{3,41E-09} \\ \hline
			L3  & \num{1,54E-07} & \num{99.85} & \num{1.51e-7} & \num{5,03E-09} \\ \hline
			L4   & \num{4,60E-07} & $>$\num{99.95} & - & \num{9,28E-07}  \\ \hline
			L5 & *\num{1.03e-6} & $>$\num{99.95} & - & -  \\ \hline
		\end{tabular}
	\end{table}

	As for Interpolation Search, since its performance is very much dependent on the data distribution, we must limit ourselves to provide an indication as to whether on datasets generated via the representative distributions included in this study, the reduction factor obtained via {\bf SLR} is such to yield a better time performance for {\bf L-IBS} with respect to {\bf IBS}. To this end, we have used the same setting as for Binary Search, except that for each table, we have computed the exact reduction factor via {\bf SLR}, rather searching for the breakeven point. In analogy with Binary Search, we have also investigated potential advantages of learned Interpolation Search on a GPU, with results that are analogous to the case of Binary Search. It is also of interest to include, for comparison,  the time performance of {\bf BFE}  and {\bf BBS} on those datasets.  The results are reported in Table \ref{T:LIBS}.  Although Learned Interpolation Search is better than its standard counterpart, due to the GPU or Binary Search results, such a fact is of methodological interest only. 
	
	\begin{table}
		\tiny
		\centering
		\caption{For each distribution, an estimate of how {\bf IBS} performs  with respect to  {\bf L-IBF}. For each column, the legend is analogous to the one in Table \ref{T:BFakeLearn}}\label{T:LIBS}
		\begin{tabular}{|c|l|c|l|l|l|}
			\hline
			\multicolumn{6}{|c|}{\textbf{Uniform}} \\ \hline \hline
			&
			\multicolumn{1}{|c|}{\textbf{ IBS}}  &
			\multicolumn{2}{c|}{\textbf{L-IBS}} &
			\multicolumn{1}{c|}{\textbf{GPU}} &
			\multicolumn{1}{c|}{\textbf{BFE}/\textbf{BBS*}} \\ \hline
			& Time (s) & \%BRF & Time (s) & Time (s) & Time (s) \\ \hline
			L1  & \num{6.56e-8} & 98.78 & \num{5.74e-8} & \num{1.81e-9} & \num{8,93E-08} \\ \hline
			L2 & \num{7.26e-8} & 99.48  & \num{6.31e-8} & \num{1.80e-9} & \num{1,13E-07} \\ \hline
			L3  & \num{1.42e-7} & 99.92 & \num{1.49e-7} & \num{1.82e-9} & \num{1,57E-07} \\ \hline
			L4  & \num{8.81e-7} & 99.99 & \num{9.57e-7} & \num{9.16e-7}  & \num{4,69E-07} \\ \hline
			L5   & \num{3.17e-6} & 99.99 & \num{2.61e-7} & - & *\num{1.03e-6} \\ \hline
			\hline
			\multicolumn{6}{|c|}{\textbf{Log-normal}} \\ \hline \hline
			&
			\multicolumn{1}{|c|}{\textbf{ IBS}}  &
			\multicolumn{2}{c|}{\textbf{L-IBS}} &
			\multicolumn{1}{c|}{\textbf{GPU}} &
			\multicolumn{1}{c|}{\textbf{BFE}/\textbf{BBS*}} \\ \hline
			& Time (s) & \%BRF & Time (s) & Time (s) & Time (s) \\ \hline
			L1  & \num{8.74e-8} & \num{64.42} & \num{1.13e-7} & \num{1.81e-9} & \num{8,46E-08} \\ \hline
			L2 & \num{1.20e-7} & \num{65.11}  & \num{3.60e-7} & \num{8.37e-9} & \num{1,08E-07} \\ \hline
			L3  & \num{3.26e-7} & \num{65.01} & \num{6.61e-7} & \num{1.82e-9} & \num{1,54E-07} \\ \hline
			L4   & \num{2.65e-4} & \num{29.67} & \num{1.26e-6} & \num{9.33e-7}  & \num{4,60E-07} \\ \hline
			L5  & \num{4.18e-4} & \num{58.25} & \num{1.49e-4} & - & *\num{1.03e-6} \\ \hline
			\hline
			\multicolumn{6}{|c|}{\textbf{Sigmoid}} \\ \hline \hline
			&
			\multicolumn{1}{|c|}{\textbf{IBS}}  &
			\multicolumn{2}{c|}{\textbf{L-IBS}} &
			\multicolumn{1}{c|}{\textbf{GPU}}  &
			\multicolumn{1}{c|}{\textbf{BFE}/\textbf{BBS*}}\\ \hline
			& Time (s) & \%BRF & Time (s)  & Time (s) & Time (s) \\ \hline
			L1  & \num{3.47e-7} & \num{82.08} & \num{1.05e-7} & \num{4.35e-9} & \num{8,52E-08} \\ \hline
			L2 & \num{5.31e-7} & \num{81.98}  & \num{1.46e-7} & \num{4.38e-9} & \num{1,12E-07}\\ \hline
			L3  & \num{1.35e-6} & \num{81.96} & \num{3.32e-7} & \num{4.41e-9} & \num{1,43E-07}\\ \hline
			L4  & \num{1.82e-5} & \num{82.08} & \num{1.36e-6} & \num{8.95e-7} & \num{4,82E-07} \\ \hline
			L5 & \num{2.19e-5} & \num{82.08} & \num{4.36e-6} & - & *\num{1.03e-6} \\ \hline
			
		\end{tabular}
	\end{table}
	
	\section{Conclusions}
	We have provided a systematic review of simple Sorted Search procedures in the setting of Learned Data Structures. Although the idea of using Machine Learning in order to speed-up Index Data Structures, for the important case of Sorted Table Search, the results are somewhat limited, in particular when a GPU is available. 
	
	\ifCLASSOPTIONcompsoc
	\section*{Acknowledgments}
	\else
	\section*{Acknowledgment}
	\fi
	Many thanks to  Giorgio Vinciguerra for helpful discusions and comments and for being so kind to run some of the experiments reported here on hardware available at his Institution. We are also grateful to Umberto Ferraro Petrillo for making available to us the Tarastat cluster at the Diparrtimento di Scienze Statistiche, Univdersit\'a di Roma \vir{La Sapienza}. 

	\ifCLASSOPTIONcaptionsoff
	\newpage
	\fi
	
	\bibliographystyle{IEEEtran}
	\bibliography{./references}

\end{document}